\documentclass[letterpaper, 10 pt, conference]{ieeeconf}  
  \usepackage{pgfplots}
  \pgfplotsset{compat=newest}
  \usetikzlibrary{plotmarks}
  \usetikzlibrary{arrows.meta}
  \usepgfplotslibrary{patchplots}
  \usepackage{grffile}
  \usepackage{amsmath}
  \usepackage{blindtext}
  \usepackage[bottom]{footmisc}
  \usepackage{mathrsfs}
  \usepackage{lipsum}
 \usepackage{url}
 \usepackage{array}
\usepackage{booktabs}
\usepackage{float}
\usepackage{multirow}
\usepackage{cite}

\newlength\fwidth
\usepackage[linesnumbered,ruled]{algorithm2e}

\usepackage{graphicx}
\usepackage{framed}
\usepackage{subcaption}
\usepackage{indentfirst}
\usepackage{multirow}
\usepackage{makecell}

\usepackage{amsmath, amssymb,amsthm}
\usepackage{color}
\usepackage{booktabs}
\usepackage{tabularx}
\usepackage{hyperref}
\usepackage[capitalise]{cleveref}
\usepackage{comment}

\IEEEoverridecommandlockouts                              

\title{\LARGE \bf
Expressing Diverse Human Driving Behavior with Probabilistic Rewards and Online Inference 
}

\author{Liting Sun$^{*}$, Zheng Wu$^{*}$, Hengbo Ma and Masayoshi Tomizuka
\thanks{*The authors are equally contributed.}
\thanks{The authors are with the Department of Mechanical Engineering, University of California, Berkeley, USA, 94720.
        {\tt\small \{litingsun, zheng\_wu, hengbo\_ma, tomizuka\}@berkeley.edu}}%
}

\begin{document}

\maketitle
\thispagestyle{empty}
\pagestyle{empty}

\begin{abstract}
In human-robot interaction (HRI) systems, such as autonomous vehicles, understanding and representing human behavior are important. Human behavior is naturally rich and diverse. Cost/reward learning, as an efficient way to learn and represent human behavior, has been successfully applied in many domains. Most of traditional inverse reinforcement learning (IRL) algorithms, however, cannot adequately capture the diversity of human behavior since they assume that all behavior in a given dataset is generated by a single cost function. In this paper, we propose a probabilistic IRL framework that directly learns a distribution of cost functions in continuous domain. Evaluations on both synthetic data and real human driving data are conducted. Both the quantitative and subjective results show that our proposed framework can better express diverse human driving behaviors, as well as extracting different driving styles that match what human participants interpret in our user study.
\end{abstract}

\section{INTRODUCTION}

\label{sec:introduction}
Understanding human behavior in the real world is important for robotic systems that interact with humans. They need to build knowledge not only at the action level, i.e., \textit{how} humans' actions change the physical states of the environment, but also at the decision level, i.e., \textit{what} humans might do given current states, and \textit{why} humans choose such actions. For example, autonomous vehicles predict future movements of other human drivers/pedestrians based on historical observations, so that they can take safe and efficient actions to transport the passengers/cargo without causing too much inconvenience to other traffic participants \cite{sun2018courteous}. Industrial collaborative robots also need prediction to help them interpret the intentions of human co-workers and provide assistance accordingly. 

To describe human behavior in human-robot interaction (HRI), particularly at the decision level, many models have been proposed in the past decades. In terms of representation, most of them can be grouped into two categories: 1) the policy representation and 2) the incentive representation. In policy representation, instead of finding out ``why humans choose such actions'', a policy is directly fit/learned connecting from ``what we have observed'' to ``what humans might do next''. Such policies can be parameterized either via pre-specified rules/models \cite{schmerling2018multimodal} or deep neural networks ( imitation learning \cite{sun2018fast, bansal2018chauffeurnet}, reinforcement learning \cite{lillicrap2015continuous, modares2015optimized, shalev2016safe, levine2018learning}). On the other hand, incentive representation tries to formulate ``why humans choose such actions'' based on the ``Theory of Mind'' \cite{hirschfeld1994mapping}. It treats humans as noisily rational optimizers whose actions are driven to pursue some internal incentives. Therefore, human behavior can be efficiently represented by cost/reward functions that capture the incentives of human actions. Such cost/reward functions are typically acquired via inverse reinforcement learning (IRL) \cite{ng2000algorithms}.

Human incentives, however, are naturally rich and diverse. For example, in a highway merging scenario, a conservative driver prefer to yield to the straight-going car, while an aggressive driver tend to pass first. Moreover, in practice, demonstrations with different cost/reward functions might be mixed in a given dataset without labeling. For instance, most of the vehicle motion datasets are recorded on public roads that contain multiple human divers with unknown but different driving preferences. Similar situations also exist for human motion datasets in HRI. A dataset might contain behavior collected from multiple people who have quite different preferences. Mixed and diverse cost functions in datasets poses a great challenge on the application of traditional IRL algorithms (\cite{ng2000algorithms, abbeel2004apprenticeship, ramachandran2007bayesian, ziebart2008maximum, levine2012continuous, wu2020efficient, Naumann2020suitability}) since they assume that all demonstrations in the training set share an identical cost/reward function. 
Some works have been proposed to relax such assumption. For example,
\cite{babes2011apprenticeship} derived an expectation-maximization (EM) approach that iteratively clusters observed trajectories and updates multiple cost functions until convergence. \cite{choi2012nonparametric} further generalized the framework by constructing a hierarchical Bayesian graph model with a non-parametric distribution over the cost function. However, the above two methods were formulated in the Markov Decision Process (MDP) setting and hard to deal with problems in large continuous domain due to the curse of dimension. Authors in \cite{nikolaidis2015efficient} addressed the problem by first clustering human demonstrations into different types via unsupervised learning based on the transition matrix between the joint human-robot actions, and then performing traditional IRL within each cluster. The features for the two steps were different, which makes it unable to assure that demonstrations clustered in the same category in the first step actually share the same reward function. 

\begin{figure*}[]
	\centering
	\includegraphics[width=0.8\textwidth]{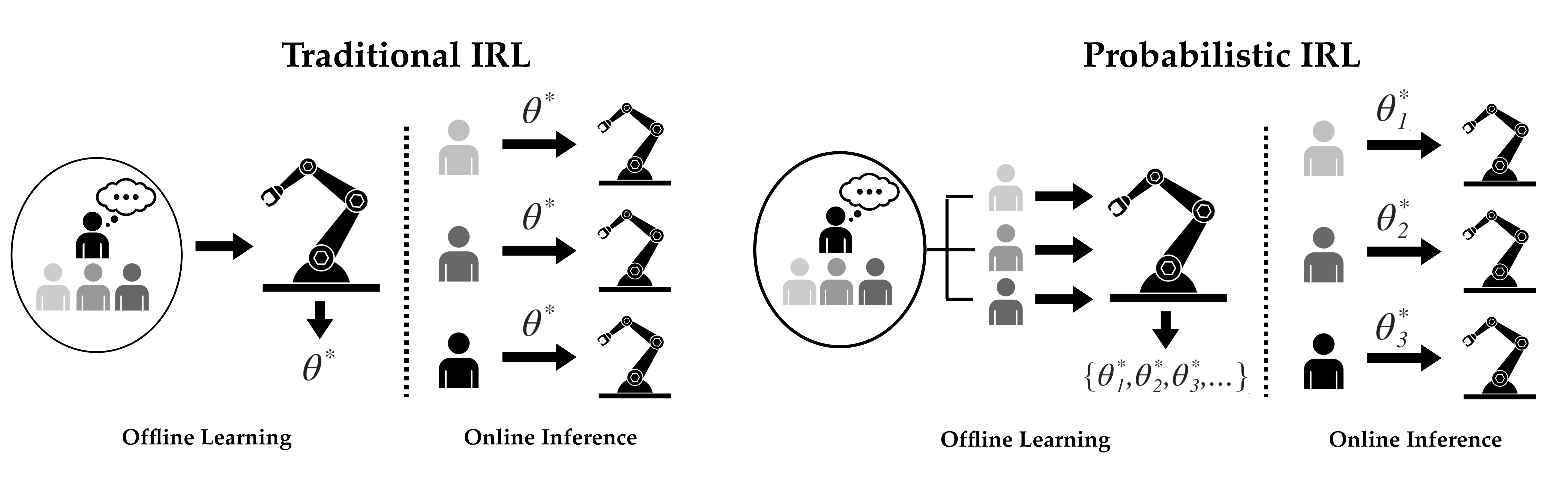}
	\caption{We propose a IRL framework to directly learn a set/distribution of cost functions from continuous-domain demonstrations with multiple unknown cost/reward functions. The traditional IRL in (a) uses a single cost function to represent behaviors from multiple agents, which fails to capture the diversity of human behavior. The proposed probabilistic IRL in (b) can extract different preferences among humans and helps improve the performance of online inference.\label{fig:contribution}}
\end{figure*}

In this work, \textit{we advocate that IRL should be able to learn a rich representation of human's incentives directly from a mixture of continuous behavior generated under multiple unknown cost/reward functions. } As shown in \cref{fig:contribution}, given a set of continuous-space demonstrations, instead of learning a single cost function to represent the average behavior (traditional IRL), we propose a probabilistic IRL that learns multiple cost/reward functions capturing the richness of human behavior. 

In summary, we make the following contributions:\\
\noindent\textbf{Developing a framework to learn multiple cost/reward functions from unlabelled data in continuous domain.} A probablistic IRL framework is developed, aiming to directly learn an optimal set or distribution of the cost/reward functions from demonstrations in continuous domain. Online inference and prediction with the multiple cost/reward functions are also addressed.

\noindent\textbf{Applying the framework to real world human driving data and extracting interpretable human driving styles}. We have applied the developed framework to real-world human driving motion data. Both quantitative and subjective tests were conducted, which showed that the framework can effectively extract different driving styles, capturing the richness of human behavior.

\noindent\textbf{Analysing the effects of prior knowledge regarding the distribution of cost functions}. We show that prior knowledge regarding the distribution of cost functions in a given set will help us gain better learn results. We implement both parametric and non-parametric prior models, and analyse the influence of prior knowledge to the algorithm performance in both offline learning and online inference.

\section{Problem Formulation}
\label{sec:problem_formulation}
In this section, the problem of continuous domain IRL from multiple agents/demonstrations with different cost/reward functions is formulated. 
Define the states of the agents as $x$, actions as $u$, and the dynamics model as 
\begin{equation}
	x_{k+1}=f\left(x_k, u_k\right).
	\label{eq:dynamics}
\end{equation}
Therefore, a trajectory contains a sequence of states and actions, i.e., $\xi=\left[x_0, u_0, x_1, u_1, \cdots, x_{N-1}, u_{N-1}\right]$, where $N$ is the length of the trajectory. 

Given a set of trajectory demonstrations $\mathbf{\mathcal{D}}$, we assume that all the agents / demonstrations share the same dynamics model in (\ref{eq:dynamics}). For instance, for human driving behavior, we assume that all the driving trajectories satisfy the kinematics of vehicle model. Moreover, we assume all agents to be noisily rational planners, i.e., their trajectories are exponentially more likely when they have lower cost. For each trajectory demonstration, the cost function is assumed to be fixed, similar to \cite{choi2012nonparametric}. Different demonstrations, however, can have different cost functions, and we do not assume further information regarding which demonstrations are generated under the same cost function and how many cost functions we might have.

We assume that cost functions can be written as linear combinations of features $\mathbf{f}(\xi)$, i.e.,
\begin{equation}
C(\xi,\theta){=}{\theta}^T\mathbf{f}(\xi) \label{eq:linear_cost}
\end{equation}
\subsection{IRL with A Single Cost Function}
Given $\mathbf{\mathcal{D}}$, traditional IRL algorithms (e.g., maximum-entropy IRL, Bayesian IRL, etc) try to find the optimal cost function, i.e., the optimal ${\theta}^*$, that maximizes the posterior likelihood of the demonstrations in $\mathbf{\mathcal{D}}$:
\begin{equation}
    {\theta}^* = \arg\max_{\theta} P(\mathcal{D}|\theta)=\arg\max_{\theta} \prod_{i=1}^{M}P(\xi_i|\theta)
    \label{eq: traditional_IRL}
\end{equation}
where $M$ is the number of demonstration trajectories in $\mathbf{\mathcal{D}}$. $P(\xi_i|\theta)$ represents the probability of trajectory $\xi_i$ given $\theta$, which can be  expressed as follows based on the principle of maximum entropy:
\begin{align}
    P(\xi_i|\theta) & \propto \exp\left(-\beta C(\xi_i, \theta)\right)\nonumber \\
    &= \dfrac{\exp\left(-\beta C(\xi_i, \theta)\right)}{\int \exp\left(-\beta C(\tilde{\xi}, \theta)\right) d\tilde{\xi}}
\label{eq:boltzman_equation}
\end{align}
$\beta$ is a hyper-parameter that reflects the rationality of the agents. From (\ref{eq: traditional_IRL}), we can see that traditional IRL algorithms inexplicitly assume that all the demonstrations in $\mathbf{\mathcal{D}}$ are generated under a single cost function which can be well represented via $\theta^*$. 

\subsection{Probabilistic IRL with Multiple Cost Functions}
There are scenarios, however, where $\mathbf{\mathcal{D}}$ contains demonstrations with multiple unknown cost functions. For example, most datasets describing human drivers' behaviour contain driving trajectories from multiple drivers who rarely share the same cost function. In such scenarios, a single optimal $\theta^*$ from (\ref{eq: traditional_IRL}) cannot well describe the diversity of behaviour. 

Instead, we assume $\theta$ as a random variable:
\begin{equation}
\theta \sim G_{\phi}
\label{eq:theta_distribution}
\end{equation}
where $G_{\phi}$ is a distribution (either in parametric or non-parametric forms) characterized by $\phi$. Each cost function in $\mathbf{\mathcal{D}}$, represented by $\theta_i$ with $i=1,2,\cdots, K{\le}M$, is an instance drawn from $G_{\phi}$. Under each $\theta_i$, the probability of trajectory $\xi_j$ is also assumed to satisfy the maximum-entropy principle as given in (\ref{eq:boltzman_equation}). $\theta_i$ is given by:
\begin{equation}
    \theta_i^\star = \arg\max_{\theta_i} P(\mathcal{D}_i|\theta_i)=\arg\max_{\theta_i} \prod_{k=1}^{M_i}P(\xi_k|\theta_i)
\end{equation}
where $\mathcal{D}_i \subseteq \mathcal{D}$, and $M_i$ is the number of trajectories in subset $\mathcal{D}_i$. During online inference, for a new demonstration, the most likely cost functions from $G_{\phi}$ to represent the observed behavior is inferred and used for prediction. We will describe more details in Section~\ref{sec: method}.



\section{Method}
\label{sec: method}
\begin{figure}[h]
	\centering
	\includegraphics[width=0.4\textwidth]{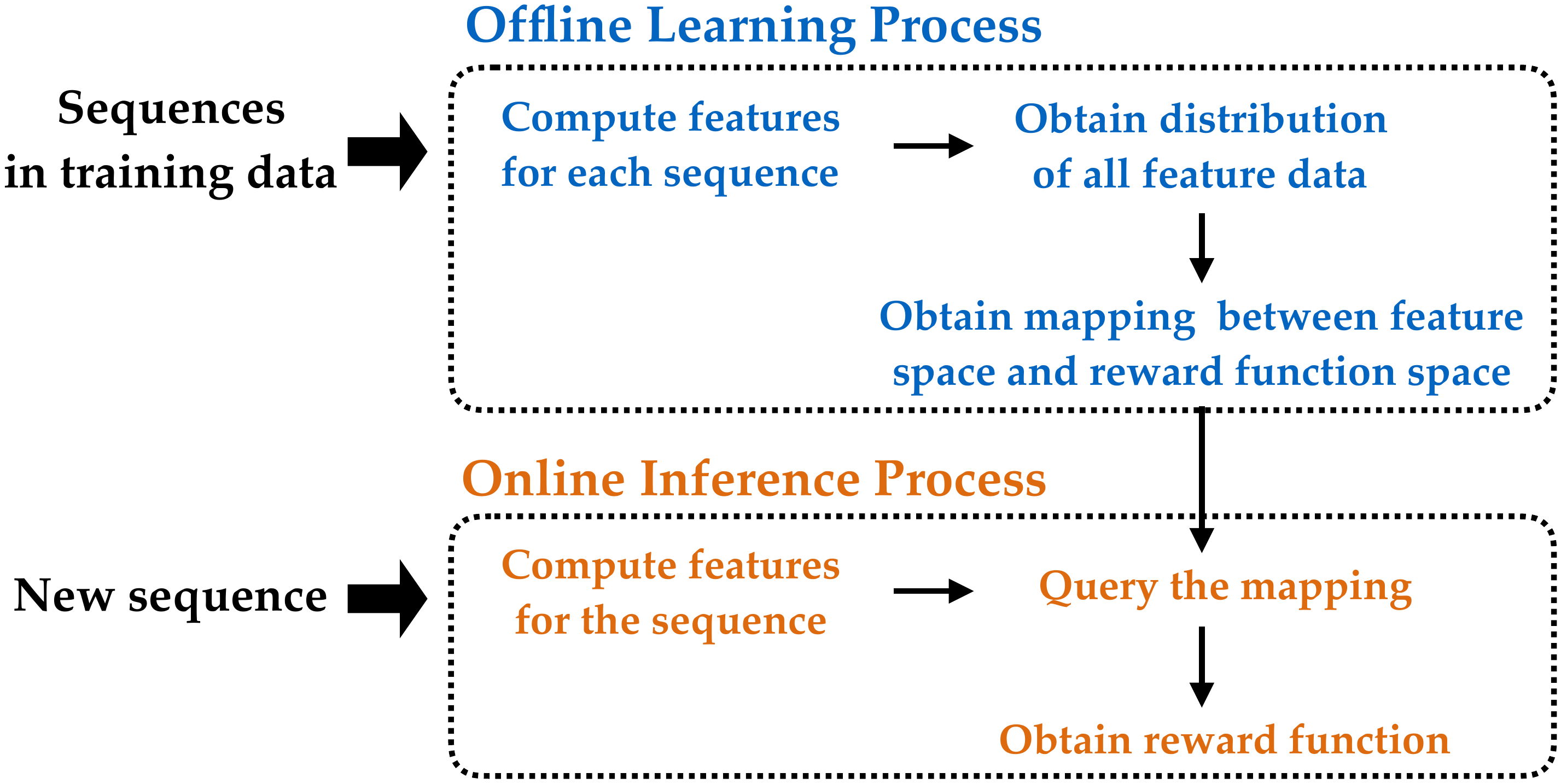}
	\caption{The pipeline of our proposed method.}
	\label{fig:algorithm_flow}
\end{figure}

To achieve the two goals for the probabilistic IRL (PIRL) in continuous domain, we propose a two-stage framework as shown in \cref{fig:algorithm_flow}. It includes an offline learning module and an online inference module. For offline learning, we first obtain a distribution over the space of cost features, i.e., the space defined by $\mathbf{f(\xi)}$ in (\ref{eq:linear_cost}), and denote it by $p(\mathbf{f(\xi)})$. Afterwards, a mapping $\mathcal{T}: p\rightarrow p$ is built to translate the distribution $p(\mathbf{f(\xi)})$ to a distribution over the cost functions, $G_{\phi}(\theta)$ as in (\ref{eq:theta_distribution}) by utilizing the IRL algorithm in \cite{levine2012continuous}.
In online inference module, for an arbitrary new demonstration data $\xi_{\text{new}}$, we query for the most likely $\theta^*$ from the learned $G_{\phi}(\theta){=}\mathcal{T}\left(p(\mathbf{f(\xi_{\text{new}})})\right)$ based on its feature $f(\xi_{\text{new}})$.

The key idea of the framework is to perform continuous-domain IRL on demonstrations with locally consistent features in feature space defined by $\mathbf{f(\xi)}$. It is based on an assumption that similarities on the feature space reflect similarities on the space of the cost functions, as suggested by the feature matching algorithm in \cite{abbeel2004apprenticeship} based on the defined linear form of cost functions in terms of features in (\ref{eq:linear_cost}). 

We start by constructing prior knowledge regarding the distribution $G_{\phi}$. In this work, we implement two types of prior distributions, one parametric and one non-parametric, as two exemplar prior models:
\begin{itemize}
	\item $G_{\phi}$ is a mixture of Gaussian distributions, i.e.,  $G_{\theta}(\phi)$ can be represented via a Gaussian Mixture Model (GMM);
	\item $G_{\phi}$ is a non-parametric distribution represented via $k$-nearest neighbors ($k$-NN).
\end{itemize}

\subsection{Prior models}
\subsubsection{GMM as a prior model}
In this scenario, we assume that the distribution over the cost functions (represented by $\theta$) can be expressed as a GMM:
\begin{equation}
G_{\phi} = \sum_{i}^{k}\omega_i P_{\phi_i}
\end{equation}
where $k$ is the number of Gaussian kernels represented by $P_{\phi_i}$ and $\omega_i$ is the weight of $P_{\phi_i}$. Within each kernel for $i{=}1,2, \cdots, k$, $\phi_i = (\mu_i,\Sigma_i)$ represent the mean $\mu_i$ and covariance $\Sigma_i$ of the kernel. Hence, $P_{\phi_i} {=} \mathcal{N}(\mu_i, \Sigma_i)$. 

With GMM as a prior, during offline learning phase, we learn the set of parameters, i.e., $(\omega_i, \phi_i), i{=}1, 2,\cdots, k$ using an EM algorithm. Each kernel encodes a cost/reward function that can be learned via the continuous-domain IRL algorithm. Among kernels, preferences vary. During online inference phase, for an arbitrary new demonstration data, we first query for the most likely kernel to obtain the most likely cost function $\theta^*$, and use $\theta^*$ to express the human behavior in the future.
\subsubsection{KNN as a prior model}
In GMM, we need to pre-define the number of the kernels $k$ which determines how many cost functions/preferences we can represent. Hence, it limits the capability of the model to express the richness of human behavior. As an extension,  we also introduce another prior: the $k$-NN model, and formulate the online inference as a $k$-NN regression problem.

Hence, during offline learning, we perform a continuous-domain IRL for each of the demonstration. During online query, we perform $k$-NN regression to find the optimal $\theta^*$ and use it to express the human behavior in the future.

\subsection{Continuous-Domain IRL}
With both prior models, we use the continuous-domain IRL proposed in \cite{levine2012continuous} to learn the cost function using either multiple demonstrations from the same kernel (GMM as a prior) or single demonstrations ($k$-NN as a prior). As mentioned in (\ref{eq:linear_cost}), we assume that cost function is parameterized as a linear combination of features, and aim to solve the problem defined in (\ref{eq: traditional_IRL}) and (\ref{eq:boltzman_equation}).

\subsection{Algorithm Implementation}
Detailed algorithms based on the two prior models and the continuous-domain IRL is given below. Algorithm \ref{alg:gmm_learning} and \ref{alg:gmm_inference} demonstrate, respectively, how the GMM model is learned offline and utilized for online inference. Algorithm \ref{alg:knn} gives the details for online inference based on a prior model with $k$-NN. Note that once we obtained the cost function, we simulate future behaviors of humans via the framework of model predictive control (MPC). Namely, at each time step, we solve an optimization problem given historial states: 
\begin{eqnarray}
\xi^*(x_0)&{=}&\arg\min_{\xi} C(x_0, \xi,\theta){=}\arg\min_{x_i, u_i} \sum_{i{=}0}^{N-1}{\theta}^T\mathbf{f}(x_i, u_i) \nonumber \\
&& \text{s.t.} \qquad x_{i+1}=f(x_i, u_i),
\label{eq:MPC}
\end{eqnarray}
where $\xi^*(x_0)$ is the predicted future trajectory. Note that at each time step $t$, we collect $T^\prime$ step historical observations and use them to infer the optimal reward/cost function. The inferred cost function is used as the input of the MPC in (\ref{eq:MPC}). Thus the inference can capture dynamically changing cost functions of humans over time, which means that if an agent change its cost function along the trajectory, the proposed algorithm can identify that and generate predictions based on the latest identified cost function. The metric $d$ in the algorithms could be chosen as Euclidean distance, i.e. $||\cdot||_2$. 

\begin{algorithm}[h]
\SetAlgoLined
\KwResult{$\{\theta_i\ , \ i=1, 2, \dots, k\}$}
 \KwIn{number of kernels of GMM $k$, initial parameters $\phi$, a metric $d$, dataset: $\mathcal{D}_n = \{f(\xi_i)\}_{i=1:n}$.}
 Get the GMM parameters $\phi$ by EM algorithm\;
 Cluster the dataset into k subsets $S_i\ , i = 1, 2, \dots, k$ according to the Gaussian mixture model\;
 Learn $\theta_i$ within each subset $S_i$ for $i=1, 2, \dots, k$ by continuous-domain IRL\;
 \caption{GMM methods\label{alg:gmm_learning}}
\end{algorithm}
\vspace{-6mm}
\begin{algorithm}[h]
\SetAlgoLined
\KwResult{$\hat{\xi}$}
 \KwIn{estimated GMM model $P_{\phi}$, a query trajectory: $q = f(\xi_{\text{new}})$}
Get the most probable cluster that the query belongs to, i.e., find out $k = \text{argmax}_{i}p_{\phi_i}(q)$ via $P_{\phi}$\;
Get the cost function $C(\xi_{\text{new}}, \theta_k) = \theta_k^T f(\xi_{\text{new}})$\;
Leverage MPC to predict the future trajectories ${\hat{\xi}_{\text{new}}}$ with the cost function $C(\xi_{\text{new}}, \theta_k) $\;
\caption{online prediction using GMM methods\label{alg:gmm_inference}}
\end{algorithm}
\vspace{-6mm}
\begin{algorithm}[h]
\SetAlgoLined
\KwResult{$\hat{\theta}(q)$}
 \KwIn{a query feature $q = f(\xi_i)$, number of nearest neighbors $k$, a metric $d$. dataset: $\mathcal{D}_n = \{f(\xi_i)\}_{i=1:n}$}
 Get the k nearest neighbors of query $q$: $\mathcal{S}_k(q) =\{s : \text{first k elements in}\ \text{sort}(d(s, q)), s \in \mathcal{D}_n\}$\;
 Calculate each $\theta(s)$ by IRL for $\forall s \in \mathcal{S}_k(q)$. with locally consistent features method\;
 $\hat{\theta}(q) \leftarrow \sum_{s \in \mathcal{S}_k(q)} \frac{1}{d(s, q)}\theta (s)$\;
 \caption{KNN methods\label{alg:knn}} 
\end{algorithm}

\vspace{-5mm}
\section{Experiments on Synthetic Data}
To verify the effectiveness of the proposed framework in \cref{sec: method}, we first conduct experiments on synthetic data.
\subsection{Synthetic Data Generation}
\begin{figure}[!h]
	\centering
	\includegraphics[width=0.3\textwidth]{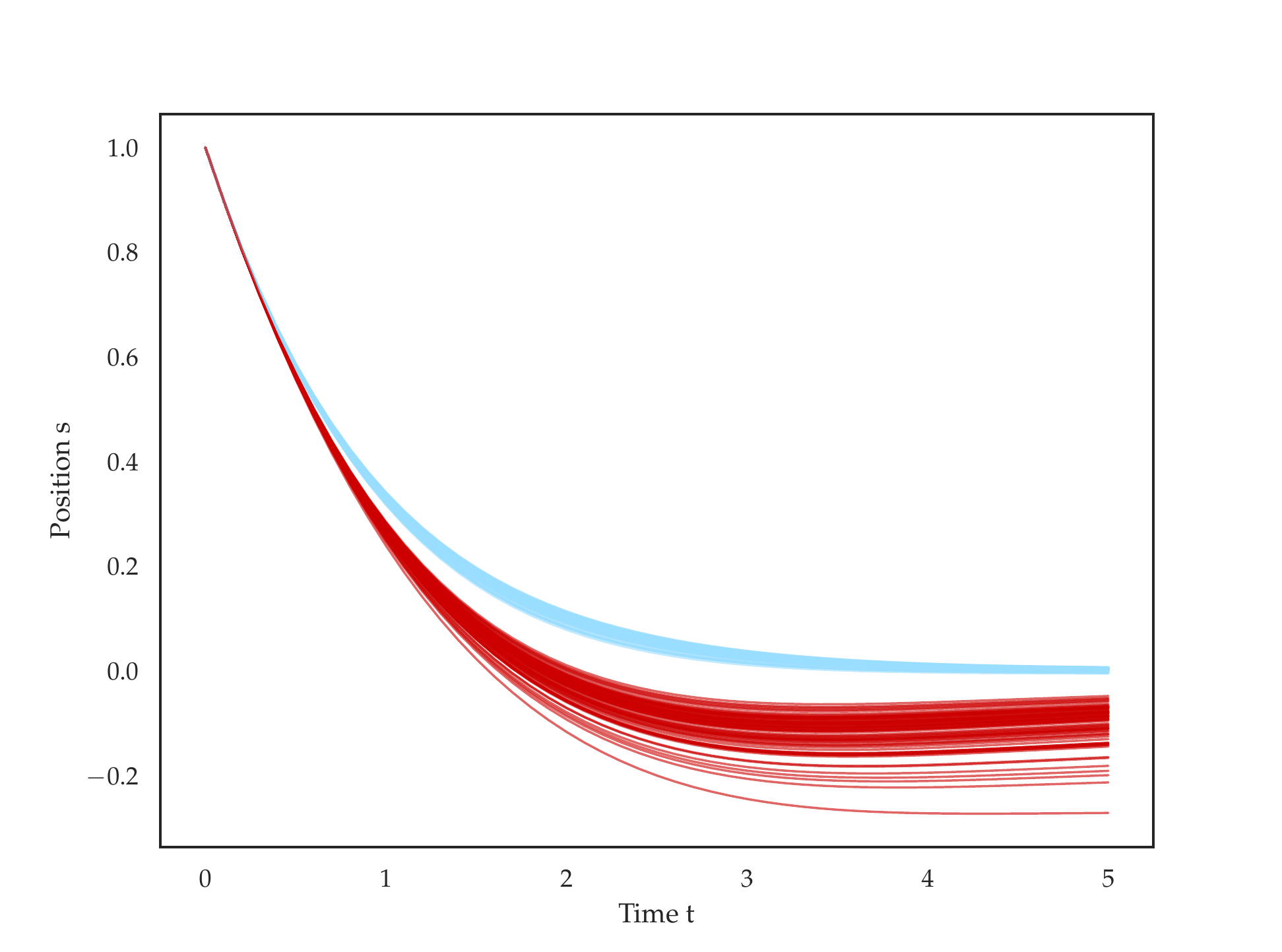}
	\caption{The synthesized trajectories generated via LQR with multiple cost functions. Red: with $q/r$ sampled from Gaussian distribution parameterized by $\sigma_1=1$, $\mu_1=0.4$; Light blue: with $q/r$ sampled from Gaussian distribution parameterized by $\mu_2=0.4$, $\sigma_2=0.1$. \label{fig:synthetic_data}}
\end{figure}

We generate our synthetic data in an optimal control problem based on a linear quadratic regulator (LQR). We let an agent track desired reference states over a finite horizon $N$. A simple point-mass kinematic model is utilized:
\begin{eqnarray}
	x_{k+1}=\left[\begin{array}{cc}
	1 & dt\\
	0 & 1
	\end{array}\right]x_{k}+\left[\begin{array}{c}
	0\\
	dt
	\end{array}\right]u_{k}\label{eq:synthetic_model}
\end{eqnarray}
where $dt$ is the sampling period. The cost function of the LQR is formulated as
\begin{equation}
J(\xi) = \sum_{i=0}^N x_i^TQx_i+ru_i^2 = r\sum_{i=0}^N x_i^T\dfrac{Q}{r}x_i+u_i^2\label{eq:lqr}
\end{equation}
with $r > 0$ and $Q \ge 0$. We assume that $Q = qI_2$ where $I_2$ is a 2-dimensional identity matrix. Hence, different preferences in LQR can be represented by the ratio $\theta=q/r$. To construct the synthetic data with multiple cost functions, we draw different samples of $\theta_i{=}q_i/r_i$ from a mixture of Gaussian distributions with two kernels represented by $\mu_1 = 4$, $\sigma_1 = 1$, $\mu_2 = 0.4$, $\sigma_2=0.1$, respectively. With each sampled $\theta_i, i{=}1, 2, \cdots, M$, we run LQR to synthesize the continuous-domain trajectories. We sampled 2,000 trajectories in total, out of which 1,800 trajectories are used as training data and 200 trajectories are used as test data. \Cref{fig:synthetic_data} shows the synthesized demonstration set, where the red trajectories are generated from $\theta$ sampled from the first kernel and the light blue one are generated via $\theta$ from the second kernel.
\vspace{-3mm}
\vspace{-5mm}
\subsection{Independent Variables}
\label{subsec:Independent_variable}
In our experiments, we controlled the methods we adopt to learn the reward functions. This independent variable takes one of three values, which we refer to as ``TIRL" (the traditional IRL), ``PIRL-GMM" (the probabilistic IRL with GMM as a prior model) and ``PRIL-kNN" (the probabilistic IRL with $k$-NN as a prior model).
\begin{enumerate}
    \item TIRL - We used traditional IRL algorithm~\cite{levine2012continuous} in training phase to learn a reward function from all demonstrations. For trajectory prediction, we solved an optimization with the learned reward function.
    \item PIRL-GMM - In the training phase, we first computed features for each demonstration in the dataset. Then all data points were clustered in the feature space via EM algorithm based on GMM and a unique reward function was learned for each cluster, as described in \cref{sec: method}. During online inference, a new data point is first clustered to find out the most probable cost function it might use, and then MPC is utilized to generate future trajectory prediction based on the corresponding cost function.
    \item PIRL-kNN - A reward function was learned for each demonstration in the training set. During online inference, $k$ nearest neighbors in the feature space were queried and $\theta$ was computed as described in~\ref{alg:knn}. The obtained cost function was used in MPC to generate the predicted trajectories. We also did experiments with different  $k$ to see how the performance varies with value of $k$.
\end{enumerate}

\subsection{Hypotheses}
\noindent\textbf{H1.    } \textit{PIRL-GMM and PIRL-kNN will be better than TIRL in capturing the diversity of cost functions, and thus perform better in trajectory prediction.} 

\subsection{Experimental Results}
To evaluate the performance of different algorithms, we compared the differences between the ground-truth trajectories in the test set and the predicted trajectories using different algorithms. We adopted Mean Euclidean Distance (MED) as a metric. It is defined as follows:
\begin{equation}
\mathcal{E} = \frac{1}{M}\sum^{M}_{i=1}{\frac{1}{N_i}\sum_{j=1}^{N_i}{\vert\xi_j^{\text{ground}} - \xi_j^{\text{prediction}}\vert}}.\label{eq:MED}
\end{equation}
$M$ is the number of trajectories in the test set, and $N_i$ is the length of each trajectory.

The results are shown in \cref{table:synthetic_result}. We can see that compared to TIRL, both PIRL-GMM and PIRL-kNN can obtain better performance. Moreover, PIRL-kNN performed better than PIRL-GMM since more local information is utilized, which increases the capability of the model in expressing more diverse behavior.
\begin{table}[!h]
	\caption{Experimental results on synthetic data (Unit: m)}
	\label{table:synthetic_result}
	\centering
	\begin{tabular}{|c|c|c|c|}
		\hline
		\multirow{2}{*}{} & \multirow{2}{*}{TIRL} & \multirow{2}{*}{PIRL-GMM (k=2)} & \multirow{2}{*}{PIRL-kNN (k=1)} \\
		&    &    &  \\ \hline
		MED & 5.6e-2   & 9.8e-3   & 3.2e-3  \\ \hline
	\end{tabular}
\end{table}

\begin{figure}[!h]
	\centering
	\includegraphics[width=0.5\textwidth]{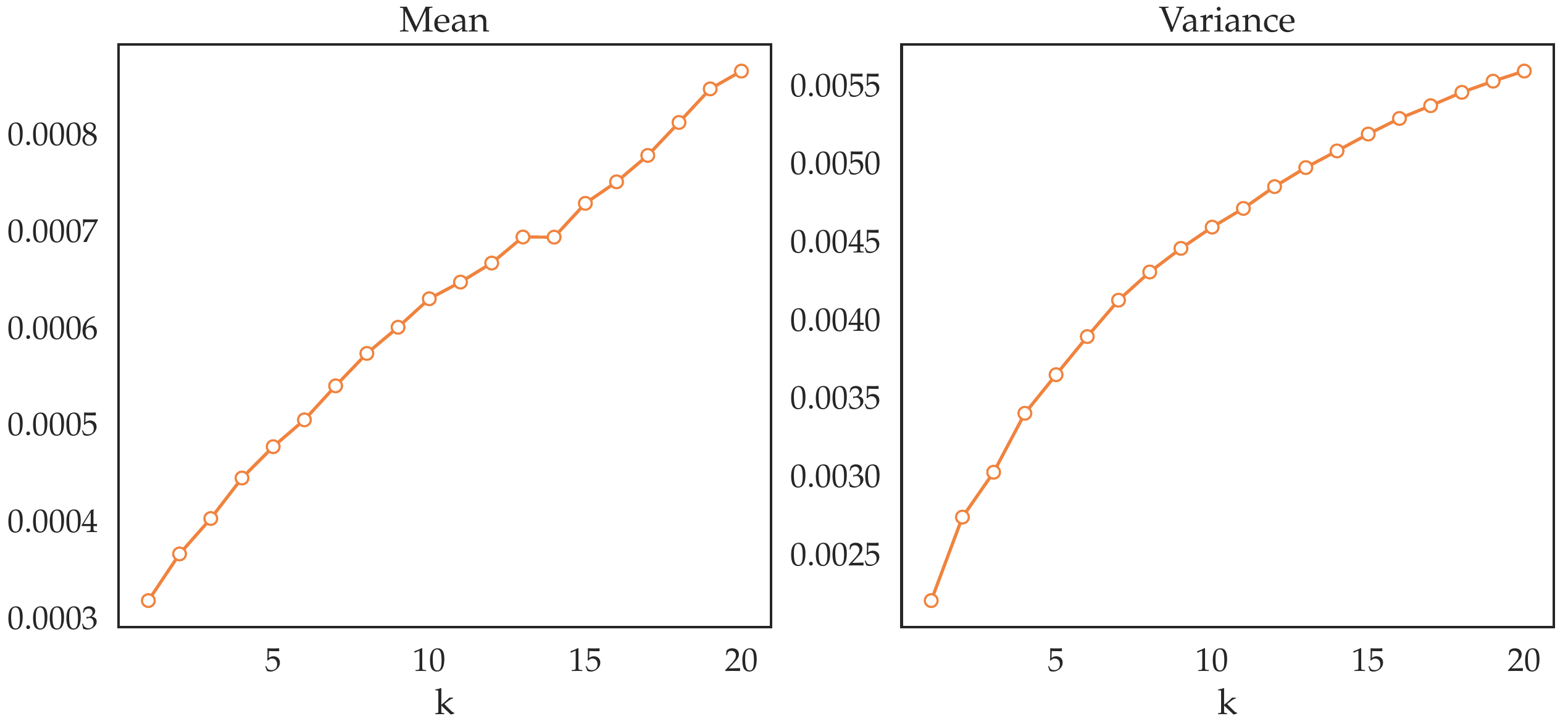}
	\caption{Performance of kNN with different $k$ on the synthetic dataset.\label{fig:knn_performance_toy}}
\end{figure}

To further investigate the influence of $k$ in the PIRL-kNN model, we tested the performance of PIRL-kNN with different $k$, and the results are given in \cref{fig:knn_performance_toy}. We can see that as $k$ increases, both the mean and variance of the prediction error increase since large $k$ reduces the capability of the model in terms of expressing more diverse behavior.

\vspace{-5mm}
\section{Experiment on Human Driving}
\label{sec:experiment_real}
We also applied the proposed framework on real human driving data to verify its effectiveness via both quantitative metrics and user studies.
\subsection{Data}
\begin{figure}[!h]
	\centering
	\includegraphics[width=0.4\textwidth]{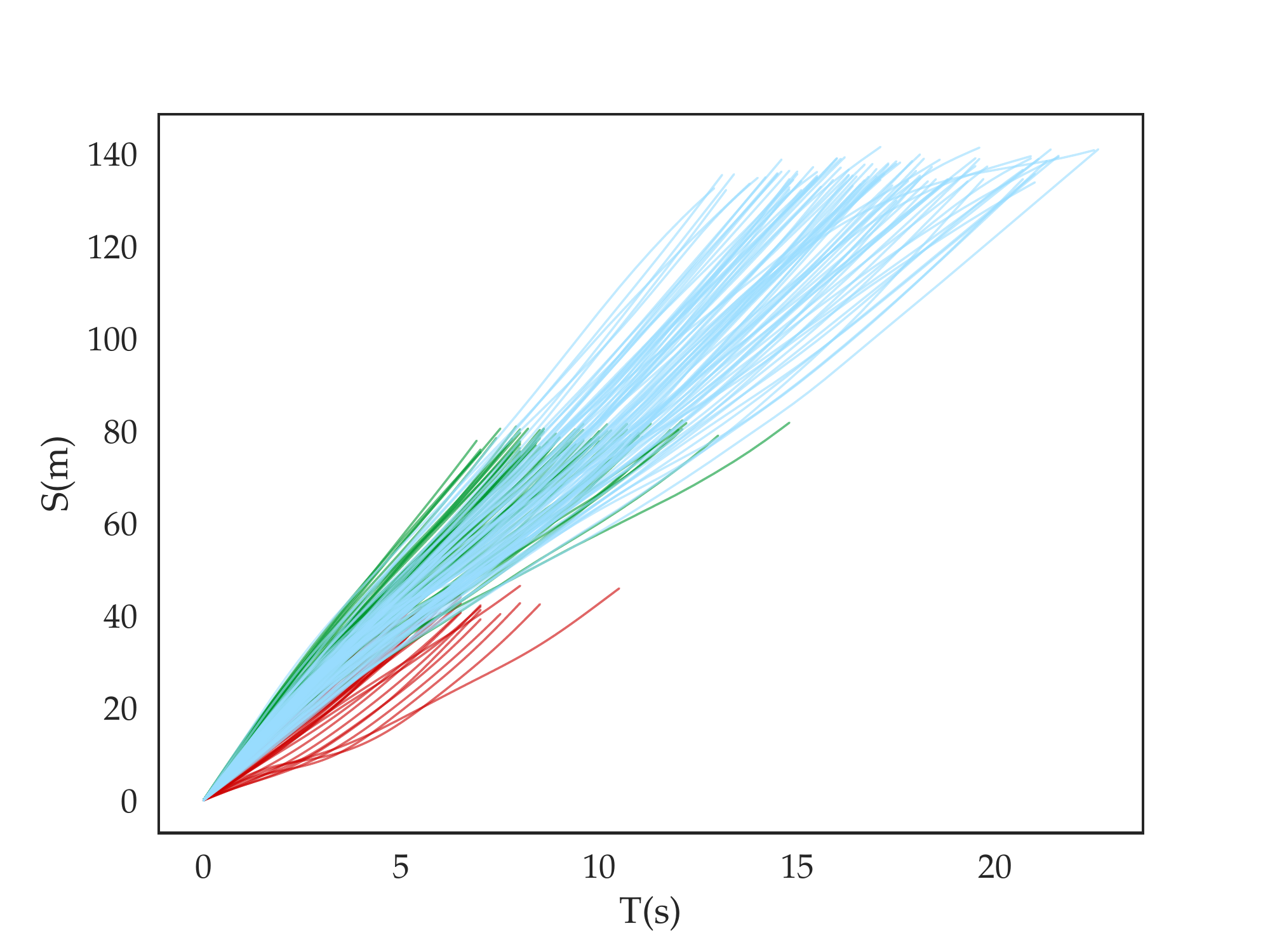}
	\caption{The selected human driving trajectories from the interaction dataset in the two roundabout scenarios.\label{fig:real_demo}}
\end{figure}

We selected the real human driving data from INTERACTION dataset \cite{zhan2019interaction} in two roundabout scenarios: DR\_DEU\_Roundabout\_OF and US DR\_USA\_Roundabout\_SR. We selected 270 driving trajectories where the drivers drove independently without too much interaction with other drivers. The sampling time of trajectory is $\Delta t {=} 0.2s$. The data is separated into two sets: a training set of size 210, and a test set of size 60. We visualized the driving trajectories in $S{-}T$ domain in \cref{fig:real_demo} ($S$: the travelled distance in meters along the reference path; $T$: the travelled time in second.). We can see that the human driving behavior is quite diverse in terms of speeds, accelerations, jerks, and so on.

\subsection{Feature Selection}
We focused on the human driving behavior in longitudinal direction, therefore we converted trajectories from Cartesian coordinate system to $S-D$ coordinate system ($S$: the travelled distance along the reference path; $D$: the deviation w.r.t. the reference path). We defined our features as
\begin{equation}
    f_1(\xi) = \frac{1}{N}\sum_{i=1}^{N}{(v_i(s)-v_{\text{desired}}(s))^2} \\
\end{equation}
\begin{equation}
    f_2(\xi) = \frac{1}{N}\sum_{i=1}^{N}{a_i^2(s)} \\
\end{equation} 
\begin{equation}
    f_3(\xi) = \frac{1}{N}\sum_{i=1}^{N}{j_i^2(s)} \\
\end{equation}
where $v_i$, $a_i$, $j_i$ denote to velocity, acceleration, jerk along longitudinal direction at time step $i$ respectively. $v_{desire}$ is calculated based on the curvature of the reference path as follows:
\begin{equation}
    v_{\text{desire}}(s) = \sqrt{a_\text{desire}\kappa(s)}
\end{equation}
where $a_\text{desire}$ is a hyperparameter and set as $1.8m/s^2$ in our experiments. $\kappa(s)$ is the curvature of the reference path.

\subsection{Independent Variables and Hypotheses}
Similar to the independent variables for synthetic data in \cref{subsec:Independent_variable}, we controlled the methods we adopt to learn the reward functions. This independent variable can take one of three values: ``TIRL", ``PIRL-GMM", and ``PRIL-kNN".

\noindent Two hypotheses are posited:

\noindent{\textbf{H2.  }} \textit{The PIRL-GMM can extract different driving styles from real human data.}

\noindent{\textbf{H3.  }} \textit{PIRL-GMM and PIRL-kNN will be better than TIRL in capturing the diversity of human behavior, and thus perform better in trajectory prediction.} 

\subsection{Experiment Results}
\begin{figure}[!h]
	\centering
	\includegraphics[width=0.4\textwidth]{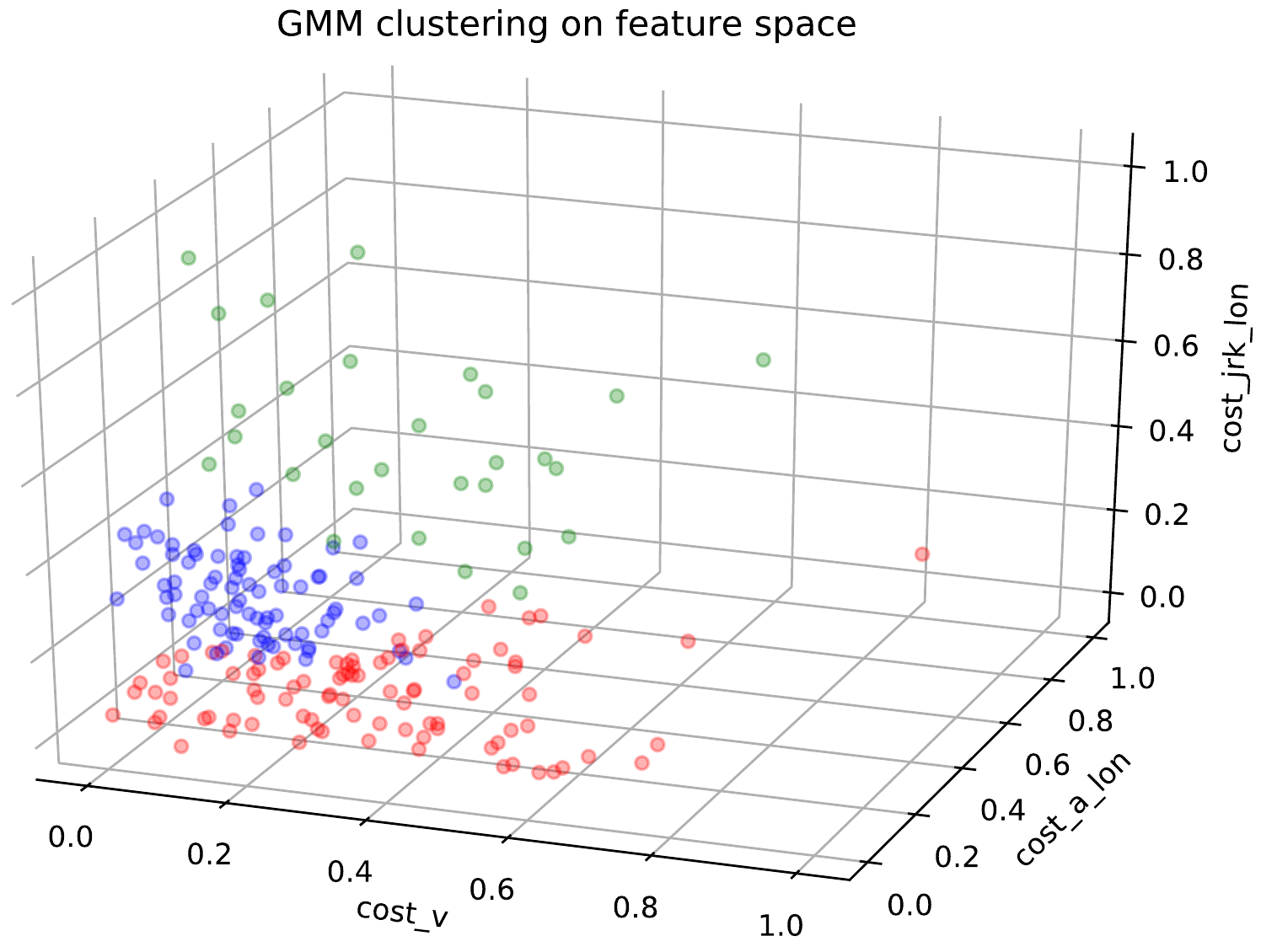}
	\caption{The clustering results in the feature space based on an EM algorithm in PIRL-GMM. Red: cluster 1; Blue: cluster 2; Green: cluster 3.\label{fig:gmm_cluster}}
\end{figure}

\Cref{fig:gmm_cluster} showed the clustering results in the feature space based on an EM algorithm in PIRL-GMM. We can see that PIRL-GMM grouped the driving trajectories into three clusters represented by red, green and blue points. In online inference, for each new data point, the PIRL-GMM will use one of the three clusters to obtain the driving style of the data. 
To evaluate the effectiveness and interpretability of such clustering results, we conducted a user study. 

\vspace{-2mm}
\subsubsection{User study I} Ten participants consisting of a mix of undergraduate and graduate students were recruited. All of the participants have technical background on autonomous driving and were comfortable and confident with questions and terminologies about different features of driving trajectories (e.g., velocity, acceleration, jerk). In the user study, each participant was shown 60 videos which contained trajectories and speed profiles generated by the cost functions learned via PIRL-GMM. The participants were asked by the following question:
\begin{itemize}
    \item ``How would you describe the driver's driving preference?".
\end{itemize}
The participants could choose an answer from three options: 1) track desired velocity, 2) minimize acceleration, and 3) no preference.

\begin{figure}[!h]
	\centering
	\includegraphics[width=0.4\textwidth]{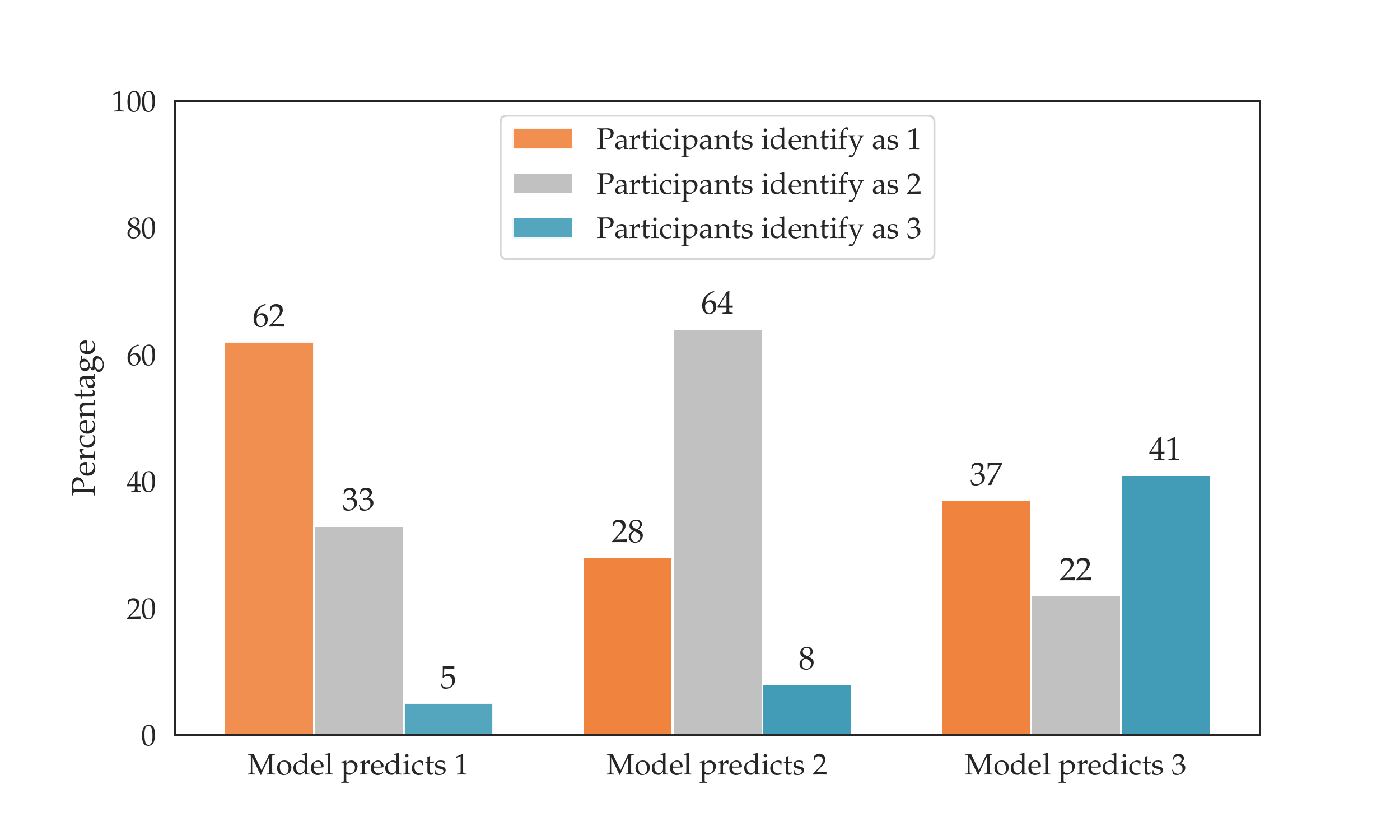}
	\caption{Results of user study I. \label{fig:result_user_I}}
\end{figure}

Results from the first user study are shown in \cref{fig:result_user_I} where the $y$-axis represents the percentages of the participants who chose the corresponding options to describe the driving styles of the tested trajectories. We can see that the identified results obtained by PIRL-GMM matched well with what participants interpreted.

\vspace{-2mm}
To prove \textbf{H3}, we evaluated the prediction performance of the three methods using both quantitative metrics and subjective metrics such as user studies. 

\subsubsection{Quantitative Results for Prediction}
Quantitatively, we again adopted MED defined in (\ref{eq:MED}) as a metric. The results are listed in \cref{tab:real_data_results}, where we can see that both PIRL-GMM and PIRL-kNN achieved better performance in prediction compared to the traditiona IRL (TIRL).
\vspace{-2mm}
\begin{table}[!h]
	\caption{MED of the prediction results on real data. $k=19$ for PIRL-kNN is shown as it is the best. (Unit: m).\label{tab:real_data_results}}
	\begin{tabular}{|c|c|c|c|}
		\hline
		\multirow{2}{*}{} & \multirow{2}{*}{TIRL} & \multirow{2}{*}{PIRL-GMM ($k=3$)} & \multirow{2}{*}{PIRL-kNN ($k=19$)} \\
		&    &    &  \\ \hline
		MED  & 9.91   & 8.67   & 6.46  \\ \hline
	\end{tabular}
\end{table}

\subsubsection{Subjective Results for Prediction: User study II}
The same ten participants were recruited in the second user study. During the user study, each participant did two sets of experiments. Within each set, each participant was shown 60 groups of videos. In each group, three videos were included, which respectively described three different driving trajectories: a ground-truth one and two simulated driving trajectories using the cost function learned via the traditional IRL and the proposed method, i.e., PIRL-GMM in experiment set I and PIRL-kNN in experiment set II. The participants were asked by the following question in each experiment set:
\begin{itemize}
    \item ``Which video do you think better mimics the driver's behavior in the real video?"
\end{itemize}

\begin{figure}[!h]
	\centering
	\includegraphics[width=0.4\textwidth]{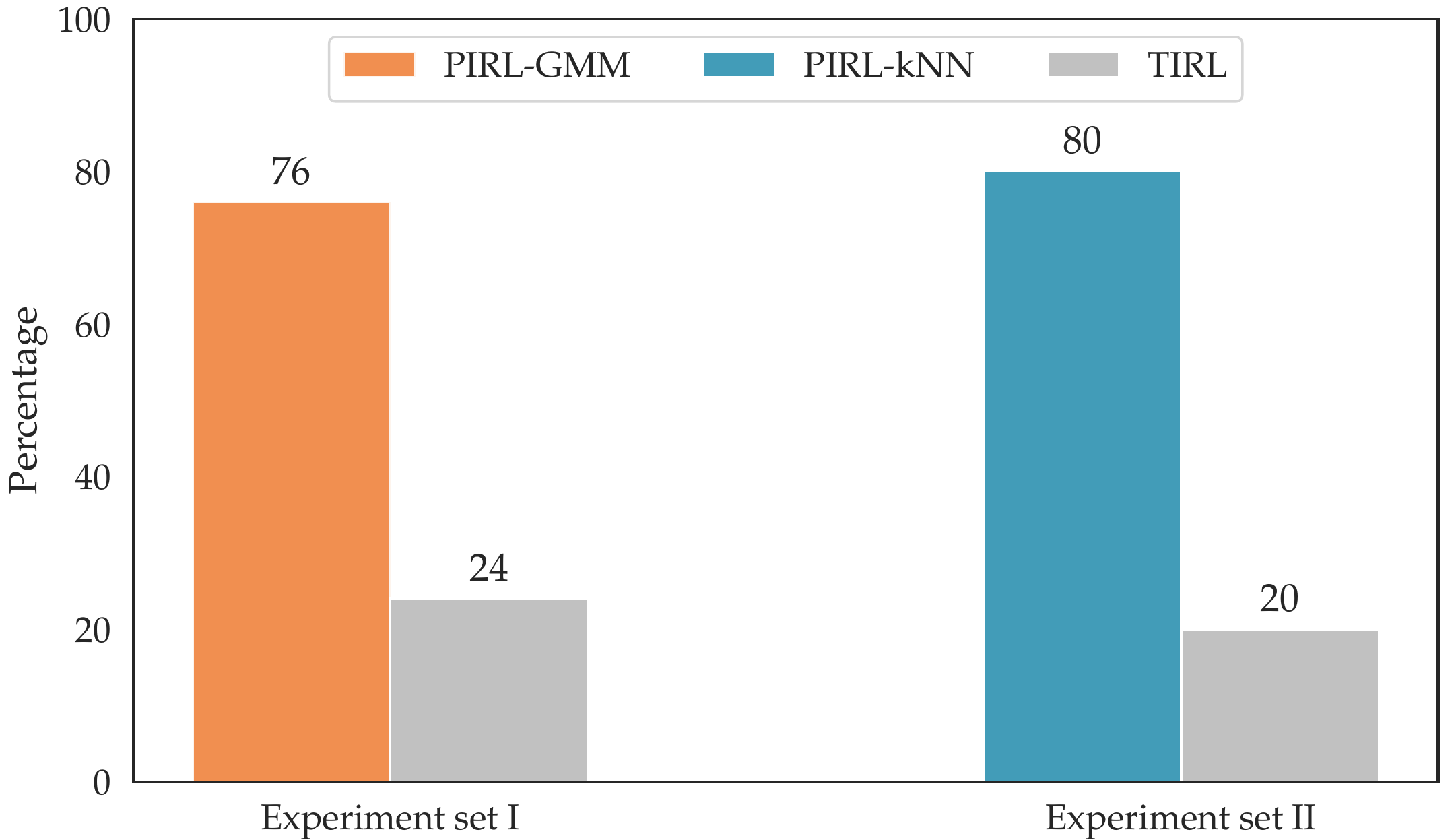}
	\caption{Results of user study II. \label{fig:result_user_II}}
\end{figure}

Results from ``User study 2" is shown in \cref{fig:result_user_II} where the $y$-axis represents the percentages of participants who believed the video generated via the corresponding method (i.e., TIRL, PIRL-GMM or PIRL-kNN) mimicked the real trajectories better. We can see that the clustering results (both the PIRL-GMM and PIRL-kNN) matched well with what participants identified. Such results also proved \textbf{H3}.

\begin{figure}[!h]
	\centering
	\includegraphics[width=0.5\textwidth]{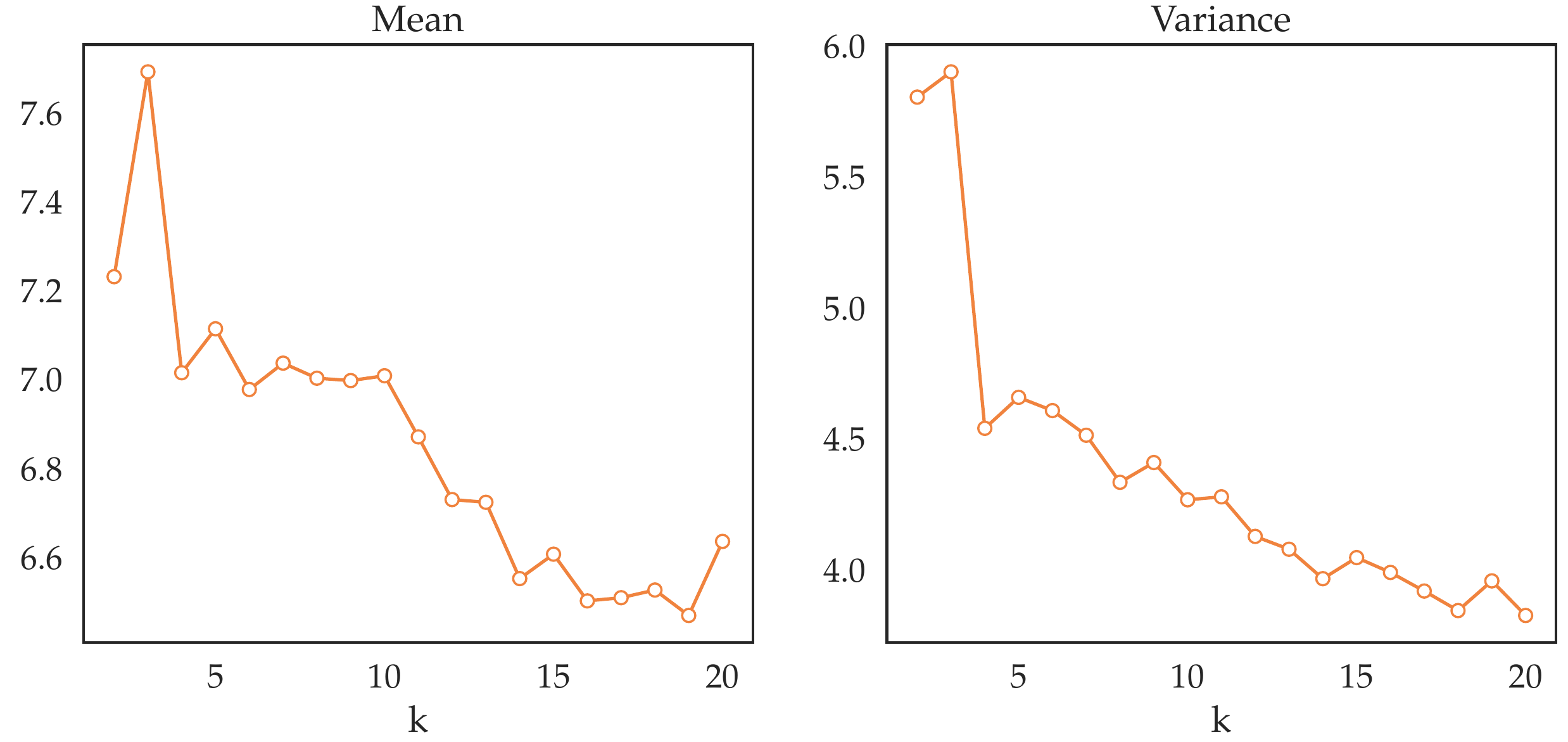}
	\caption{The performance of PIRL-kNN on real dataset as $k$ increases.\label{fig:knn_real}}
\end{figure}

We also have investigated the influence of $k$ to the performance of PIRL-kNN using real data. As we were varying $k$, the results of PIRL-kNN are shown in \cref{fig:knn_real}. Different from the results in synthetic data shown in \cref{fig:knn_performance_toy}, as $k$ increases, the performance of $k$-NN on real data also increases with reducing prediction errors (both the mean and variance). The reason behind such controversial results is that the real data is contaminated by measurement noises and quite sparse. Therefore, although using only local information in the feature space to query for the cost functions can enhance the expressive capability of the model in terms of diversity, the model also suffers from poor robustness. Hence, when we apply the PIRL-kNN to real world data, careful attention should be paid in selecting an appropriate hyper-parameter $k$ to balance the expressive capability and the robustness of the model.

\vspace{-6mm}
\section{Conclusion}
\label{sec: conclusion}
In this paper, we proposed a probabilistic IRL framework which can directly learn a distribution of cost function from multiple demonstrations with different cost functions. We considered two prior models, i.e., GMM and $k$-NN, and developed corresponding algorithms for offline learning and online inference. Two sets of experiments were conducted, with one on synthetic data and one on real human driving data. The results from both experiments verified the effectiveness of the proposed framework: it can better capture the diversity of human behavior and thus achieve better prediction performance. Moreover, results from user studies also verified that the proposed framework can extract human-interpretable driving styles.

The work in this paper can be further extended in many directions. One direction is to increase the expressiveness of cost functions. We used a linear combination of manually selected features as the cost functions, however, human incentives might not be only compromise among such features, but contains other functional formats. We will explore humans' diversity at that level in future works.
\label{sec:conclusion}



\bibliographystyle{IEEEtran}
\bibliography{references}

\end{document}